\documentclass[10pt,twocolumn,letterpaper]{article}

\usepackage[pagenumbers]{cvpr} % To force page numbers, e.g. for an arXiv version

\usepackage{multirow}
\usepackage{booktabs}
\usepackage{siunitx}
\usepackage{pifont}
\usepackage[accsupp]{axessibility}  % Improves PDF readability for those with disabilities.
\definecolor{cvprblue}{rgb}{0.21,0.49,0.74}
\usepackage[pagebackref,breaklinks,colorlinks,allcolors=cvprblue]{hyperref}
\newlist{myitemize}{itemize}{1}
\setlist[myitemize,1]{label=\textbullet,leftmargin=5.5mm}
\def\paperID{12070} % *** Enter the Paper ID here
\def\confName{CVPR}
\def\confYear{2026}

\title{Generative Video Compression with One-Dimensional Latent Representation}

\author{Zihan Zheng$^1\thanks{ This work was done when Zihan Zheng, Zhaoyang Jia and Naifu Xue were full-time interns at Microsoft Research Asia. }\quad\!$ Zhaoyang Jia$^{1*}\quad\!$ Naifu Xue$^{2*}\quad\!$ Jiahao Li$^3\quad\!$ Bin Li$^3 \quad\!$ Zongyu Guo$^3 \quad\!$\\ Xiaoyi Zhang$^3 \quad\!$ Zhenghao Chen$^4 \quad\!$ Houqiang Li$^1 \quad\!$ Yan Lu$^3 $ \\
$^1$ University of Science and Technology of China $\ \ ^2$ Communication University of China \\ $\ \ ^3$ Microsoft Research Asia $\ \ ^4$ University of Newcastle\\
{\tt\small \{zzh2003, jzy\_ustc\}@mail.ustc.edu.cn, lihq@ustc.edu.cn}
{\tt\small , zhenghao.chen@newcastle.edu.au}\\
{\tt\small \{li.jiahao, libin, zongyuguo, xiaoyizhang, yanlu\}@microsoft.com}
{\tt\small , aaronxuenf@cuc.edu.cn}
}
\vspace{-5mm}
\begin{document}
\maketitle
\begin{abstract}
Recent advancements in generative video codec (GVC) typically encode video into a 2D latent grid and employ high-capacity generative decoders for reconstruction. However, this paradigm still leaves two key challenges in fully exploiting spatial-temporal redundancy:
Spatially, the 2D latent grid inevitably preserves intra-frame redundancy due to its rigid structure, where adjacent patches remain highly similar, thereby necessitating a higher bitrate.
Temporally, the 2D latent grid is less effective for modeling long-term correlations in a compact and semantically coherent manner, as it hinders the aggregation of common contents across frames.
To address these limitations, we introduce \textbf{G}enerative \textbf{V}ideo \textbf{C}ompression with One-Dimensional (\textbf{1D}) Latent Representation (GVC1D). GVC1D encodes the video data into extreme compact 1D latent tokens conditioned on both short- and long-term contexts. Without the rigid 2D spatial correspondence, these 1D latent tokens can adaptively attend to semantic regions and naturally facilitate token reduction, thereby reducing spatial redundancy. Furthermore, the proposed 1D memory provides semantically rich long-term context while maintaining low computational cost, thereby further reducing temporal redundancy.
Experimental results indicate that GVC1D attains superior compression efficiency, where it achieves bitrate reductions of 60.4\% under LPIPS and 68.8\% under DISTS on the HEVC Class B dataset, surpassing the previous video compression methods.
Project: \url{https://gvc1d.github.io/}
\end{abstract}
\section{Introduction}
\label{sec:intro}

\begin{figure}[t]
% \vspace{-2mm}
    \centering
    \includegraphics[width=1\linewidth]{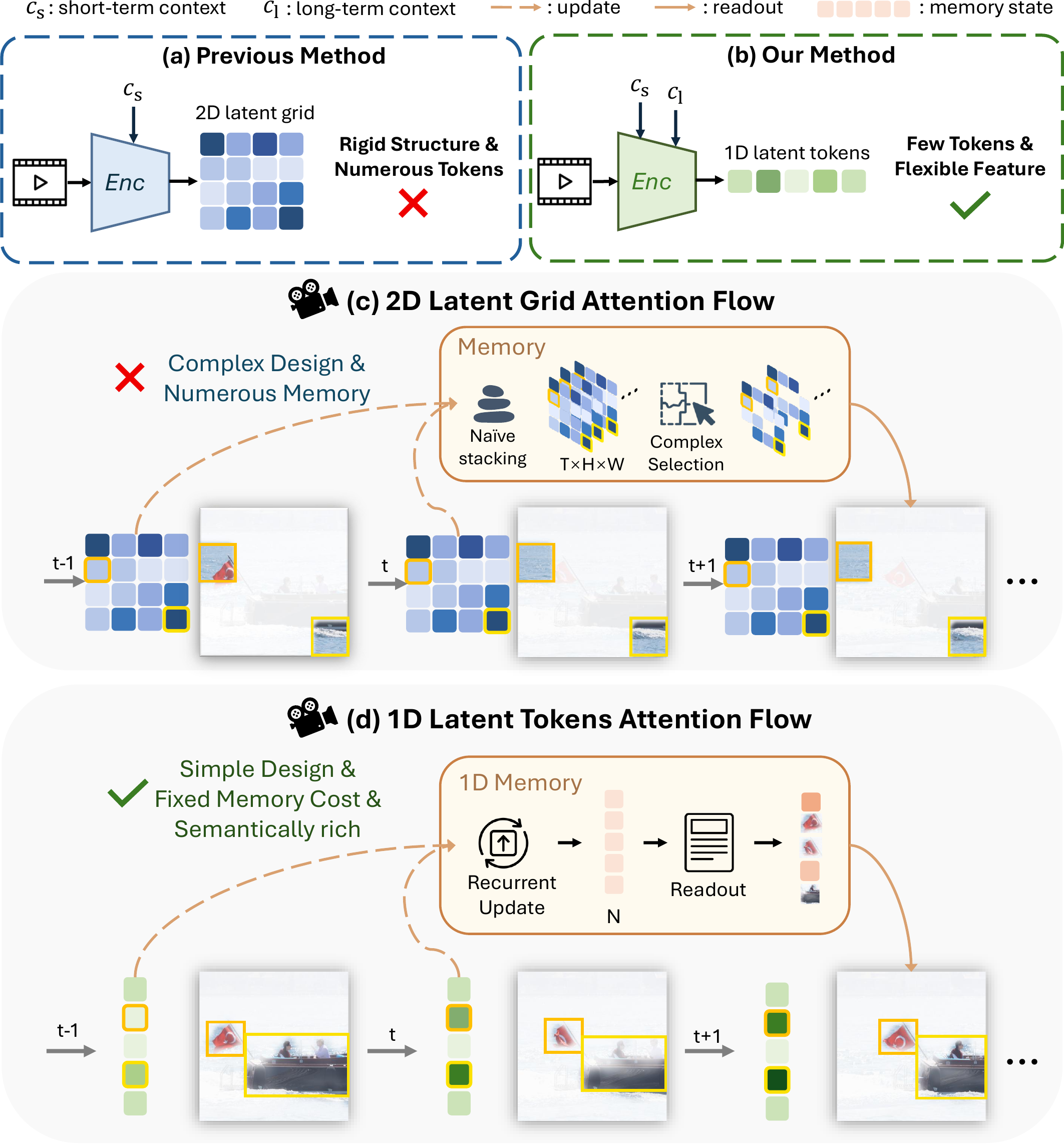}
      \vspace{-4mm}
    \caption{\textbf{Method comparison}. (a) Previous generative video codecs~\cite{qi2025generative,ma2025diffusion,yang2022perceptual} encode videos into dense 2D latent grids with rigid spatial structures using short-term context $c_{s}$, resulting in numerous inflexible tokens. (b) Our method exploits short-term $c_{s}$ and 1D-based long-term context $c_{l}$ to encode videos into a few flexible 1D latent tokens. \textbf{Attention map comparison}. (c) 2D latent grids preserve fixed spatial correspondences between tokens and image patches, limiting redundancy exploitation and requiring complex memory designs~\cite{wang20243d, VTM, qian2024streaming}. (d) Our 1D latent tokens adaptively attend to semantic regions, while the 1D memory, managed by a few 1D tokens, efficiently preserves long-term context in a semantically coherent and computationally efficient manner.}
    \label{fig:intro_0}
     \vspace{-6mm}
\end{figure}

As video data continues to grow, achieving efficient compression while preserving human perceptual quality has become critical for transmitting high-fidelity content at low bitrates in applications such as video streaming and conferencing. However, both traditional codecs~\cite{HM, VTM, ECM} and recent neural video codecs (NVCs)~\cite{li2024neural, jia2025towards, tang2025neural}, which are typically optimized for generic distortion measures, often yield blurry reconstructions due to limited preserved information~\cite{blau2019rethinking}.
To improve perceptual quality, recent generative video codecs (GVCs)~\cite{qi2025generative,ma2025diffusion,yang2022perceptual,xue2025single} encode raw video into latent representations (typically 2D feature maps) and decode them with powerful generative models~\cite{rombach2022high,van2017neural} to add fine details. Coupled with perceptual loss, these GVCs achieve better perceptual performance.
Regardless of their architectural designs or optimization objectives (distortion- or perception-oriented), both traditional and neural approaches ultimately transform the raw signal into 2D spatial structures, such as blocks in traditional codecs or 2D feature maps in neural ones, for subsequent processing such as entropy coding and decoding.

Although 2D representations are intuitive and naturally align with the spatial structure of images, representing frames as 2D latent grids introduces several limitations.
Firstly, each token in a 2D latent grid models only its corresponding image patch and focuses more on spatial modeling than object semantics.
They capture only variations within its local region, without understanding how the object moves as a whole.
Secondly, the 2D latent grid is dense and redundant due to its rigid spatial structure and additional spatial dimension: each image patch is encoded into a token of equal size, resulting in an excessive number of tokens. For instance, both complex and simple regions are assigned the same number of tokens, introducing unnecessary redundancy. 
These limitations stem from the 2D latent grid’s preservation of spatial structure, which enforces a rigid mapping between 2D latent grids and image patches, thereby restricting the model’s capacity to exploit image redundancy for a more compact latent representation.

To overcome these limitations imposed by 2D structures, we propose compressing videos into a 1D latent representation.
Notably, 1D latent representations have been shown to capture high-level semantics efficiently and yield more compact latent spaces in image processing tasks~\cite{yu2024an, yu2024randomized, kim2025democratizing, miwa2025one}. In particular for image compression~\cite{jia2025cod,xue2025dlf,xue2025one}, prior work~\cite{xue2025dlf} has encoded images into 1D latent sequences, achieving extremely low bitrates while preserving competitive perceptual performance.
In this work, we employ a Vision Transformer–based 1D tokenization to encode videos into 1D latent tokens, which serve as the 1D latent representation.
Specifically, our 1D representation refers strictly to the compressed latent bottleneck. While 2D spatial information is initially processed, we use a set of unstructured, learnable 1D tokens to aggregate information from 2D patch embeddings via a ViT, rather than simply flattening them into a sequence. 
These tokens effectively aggregate essential information from 2D patches without enforcing fixed spatial correspondences (as illustrated in Fig.~\ref{fig:intro_0}), freeing them from the rigidity and locality constraints of 2D grids. This design decouples the number of tokens from the spatial resolution, contributing to a compact and semantically rich latent space that enables more flexible modeling of semantic variations and improves compression efficiency.
Furthermore, by removing rigid spatial correspondences and eliminating the additional spatial dimension, the 1D latent tokens facilitates token reduction, allowing a compact set of latents to efficiently represent video frames and thereby lowering the bitrate.

Beyond effectively removing spatial redundancy, the proposed 1D latent representation enhances long-range dependency modeling, thereby strengthening conditional video compression.
Prior studies~\cite{qi2024long, phung2025mh} have demonstrated the benefit of modeling long-term correlations for conditional coding. Yet, their reliance on 2D latent representations limits the ability to provide temporally consistent semantic features. These representations primarily capture spatial variations across frames rather than semantic dynamics, which ultimately constrains long-term context preservation.
Hence, leveraging its compactness and enhanced semantic feature modeling, as illustrated in Fig.~\ref{fig:intro_0}, we further propose a 1D memory that recurrently updates a fixed-size memory state with a small set of 1D latent tokens. 
Owing to the semantically rich input, the 1D memory can extract temporally consistent semantic features without being constrained by spatial structures, thereby providing a semantically rich and coherent context. Moreover, due to the compactness of the 1D input, the 1D memory can store richer semantic contextual information within the same memory size while keeping the computational cost low. Benefiting from these properties, our memory adopts a simple yet effective design (Transformer) for long-term context modeling. The proposed 1D memory captures abstract long-term semantics through a small number of semantically rich 1D tokens, while the features from the preceding frame preserve fine-grained structural details. Together, they provide a more comprehensive context and further reduce redundancy.

To the best of our knowledge, this work is the first to employ a 1D latent representation for video compression. Integrating our design of 1D latent tokens and the 1D memory, we propose \textbf{G}enerative \textbf{V}ideo \textbf{C}ompression with One-Dimensional (\textbf{1D}) Latent Representation (GVC1D), demonstrating that the 1D latent representation achieves better performance by removing the rigid spatial structure of 2D representations. Combined with our long-term context constructed from a few 1D tokens, our method achieves state-of-the-art (SOTA) compression performance across multiple datasets and metrics, demonstrating particularly strong performance. Compared to the previous perceptual codec~\cite{qi2025generative}, our method reduces the bitrate by 60.4\% in LPIPS and 68.8\% in DISTS on the HEVC-B dataset. Furthermore, our approach provides visually superior results relative to existing video codecs~\cite{ECM, qi2025generative, li2024neural}.

In summary, our contributions are as follows:
\begin{itemize}
    \item We propose a novel video compression method based on 1D latent representations. By removing the rigid 2D spatial structure and reducing the number of tokens, our model encodes videos into a few flexible 1D tokens, enabling more efficient and adaptive video compression. 
    \item We introduce a long-term memory module that leverages 1D latent tokens to construct semantically rich context, enhancing conditional coding through improved contextual modeling.
    \item Our approach achieves state-of-the-art performance across multiple benchmarks, reducing bitrate while maintaining high perceptual and visual quality.
\end{itemize}

\section{Related Work} 
\label{sec:related_work}

\vspace{1mm}
\noindent \textbf{Learned Video Compression.}
Learned video codecs achieve compression by leveraging neural networks to remove redundancy. 
For neural video codecs (NVCs) primarily optimized for PSNR, the DCVC series~\cite{li2021deep,sheng2022temporal,li2022hybrid,li2023neural,li2024neural,jia2025towards} has demonstrated remarkable potential by replacing traditional residual coding with conditional coding. Through continuous architectural improvements, the latest DCVC-RT~\cite{jia2025towards} even surpasses advanced traditional codecs~\cite{HM, VTM, ECM} while maintaining high coding speed. However, since these models are mainly optimized for PSNR, they often suffer from severe detail loss at low bitrates, producing blurry frames. Consequently, improving perceptual quality has become a crucial research direction.
For generative video codecs (GVCs), the goal is to produce more realistic results using powerful generative models. For example, GLC-Video~\cite{jia2024generative,qi2025generative} achieves high perceptual compression quality at low bitrates by performing transform coding in the 2D latent space of a vector-quantized variational autoencoder (VQ-VAE)~\cite{van2017neural}, which is better aligned with human perception but suffers from limited performance due to the restricted capacity of VQ-VAE. Other models, such as DiffVC~\cite{ma2025diffusion}, employ more powerful pretrained diffusion models~\cite{rombach2022high} to enhance perceptual quality; however, their bitrates remain relatively high, failing to fully demonstrate the advantages of GVCs at low bitrates.
Our method also focuses on improving perceptual quality. Unlike previous approaches that primarily enhance codec modules while overlooking the critical role of token reduction and token structure in compression efficiency, we address this challenge from a representational perspective by employing a small number of flexible and semantically rich 1D latent representations to improve perceptual quality.

\vspace{1mm}
\noindent \textbf{1D Tokenization.}
In recent years, 1D image tokenization has achieved remarkable progress in image generation~\cite{yu2024an, yu2024randomized, kim2025democratizing, miwa2025one}, enabling the synthesis of high quality images with lots of details. TiTok~\cite{yu2024an} demonstrated the potential of discrete 1D latent tokens as a replacement for the 2D latent grid, achieving more compact semantic compression that enhances generation quality, while reducing the number of tokens to improve generation speed. TA-TiTok~\cite{kim2025democratizing} extended discrete 1D latent tokens to continuous formats and incorporated textual information, thereby enhancing alignment with text captions. One-D-Piece~\cite{miwa2025one} further advanced this direction by employing the Tail Token Drop regularization technique to develop a variable-length discrete image tokenizer.
In the field of image compression, DLF~\cite{xue2025dlf} employs discrete 1D tokenization~\cite{yu2024an} to extract compact semantic representations, achieving superior compression performance through a semantic branch. However, this discrete 1D tokenization is unsuitable for video compression, as the discrete format undermines the video temporal consistency.
Moreover, most existing video compression methods still rely on 2D representation. Yet, the dense structure and rigid spatial constraints of 2D representation restrict their ability to achieve efficient compression. Motivated by the continuous 1D tokenizers introduced in TA-TiTok~\cite{kim2025democratizing}, we enhance video codec efficiency by reducing the number of tokens and exploiting the compact semantic representations offered by 1D latent tokens.

\vspace{1mm}
\noindent \textbf{Long-term Context.}
Long-term context has been extensively leveraged in various tasks, such as video understanding~\cite{di2025streaming, qian2024streaming, he2024ma}, to capture long-range dependencies.
Some approaches~\cite{wang20243d, wang2025continuous} use fixed-size or bounded dynamic memory to preserve long-term context but suffer from information loss due to dense input features and limited memory capacity. In contrast, other methods~\cite{hesham2025exploiting, zatsarynna2025manta} reduce computational overhead by adopting suboptimal state-space models (SSMs), thereby sacrificing the strong representational capability of Transformers.
In video compression, DCVC-LCG~\cite{qi2024long} introduces a complex intra-cluster search mechanism based on dense 2D features to capture long-term context.
In contrast, our method manages the 1D memory using only a few semantically rich 1D latent tokens. The compact input enables the use of a full Transformer while maintaining low computational cost and mitigating information forgetting. Furthermore, the semantic richness of these 1D tokens allows the memory to naturally focus on compact and meaningful representations without requiring additional complex designs.

\begin{figure*}[t]
    \begin{center}
        \includegraphics[width=1.0\linewidth]{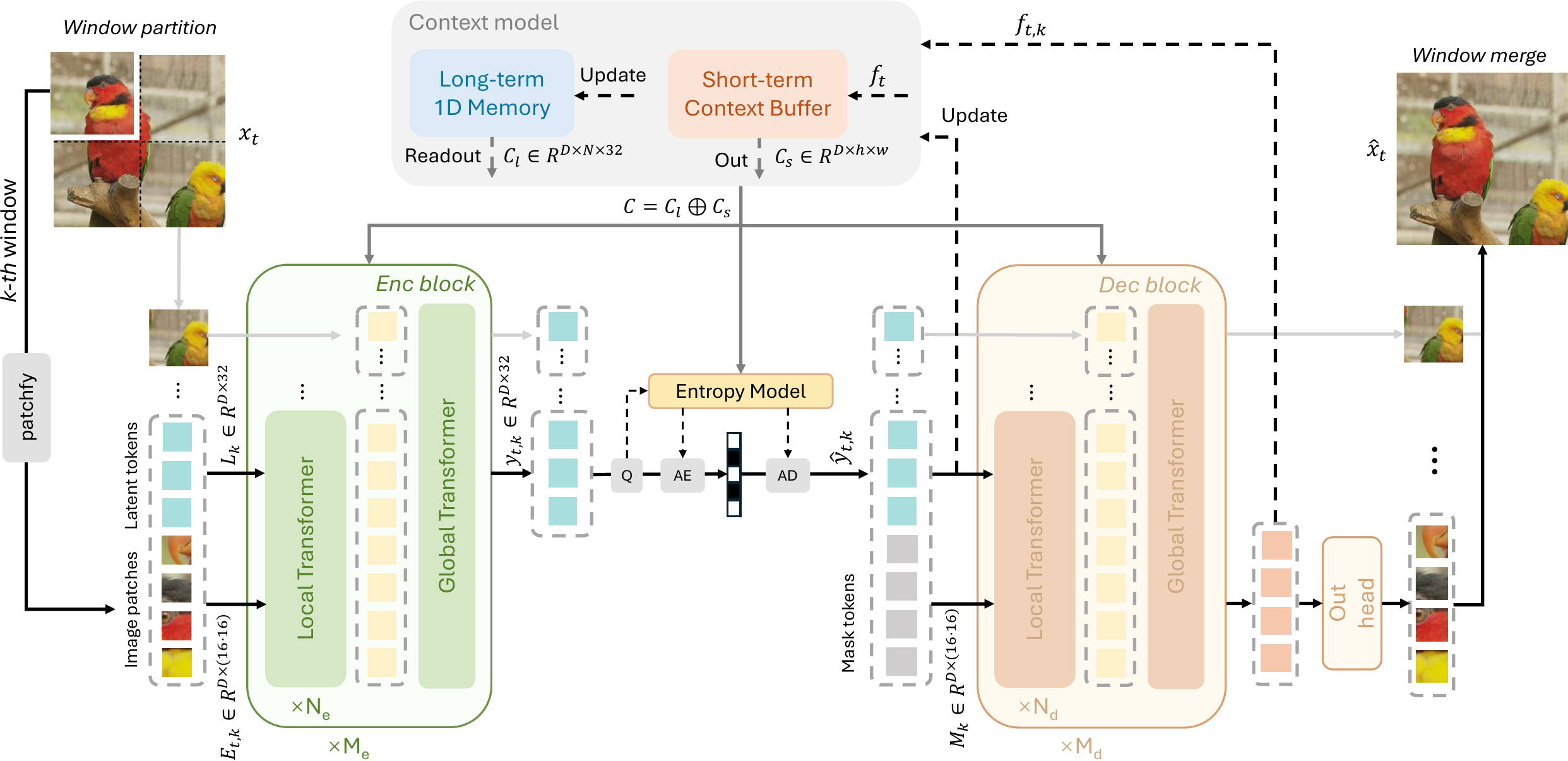}
    \end{center}
    \vspace{-6mm}
    \caption{
        Framework overview. Q, AE and AD represent quantization, arithmetic encoder and decoder, respectively. The input image $x_t$ is first embedded into patches and then fed into the encoder composed of local and global transformers to produce $y_t$. The local transformer handles each window independently with low cost, while the global transformer captures inter-window correlations. Subsequently, an entropy model then performs autoregressive entropy coding on $y_t$. The decoder adopts an architecture similar to the encoder to reconstruct the image $\hat{x}_t$. All are guided by a context model combining long-term 1D memory and short-term context buffer to provide comprehensive temporal context.
    }
    \label{fig:method_1}
    \vspace{-5mm}
\end{figure*}
\section{Methodology}
\label{sec:method}

\subsection{Framework Overview}
\label{sec:method-Pipeline }
\noindent Existing generative video codecs~\cite{qi2025generative,ma2025diffusion,yang2022perceptual} typically encode the current frame $x_t$ into a dense 2D latent grid using the short-term context from the preceding frame. However, this approach constrains their ability to effectively exploit the redundancy inherent in videos due to the rigid spatial structure of the 2D latent grid and the inefficient use of temporal information. In contrast, our method encodes $x_t$ into a few 1D latent tokens by leveraging both the short- and long-term context. As illustrated in Fig. \ref{fig:method_1}, we first employ a Vision Transformer-based~\cite{dosovitskiy2020vit} encoder to encode the current frame $x_t$ into a few 1D latent tokens $y_t$. Next, we perform entropy coding in these 1D latent tokens to achieve efficient compression. Finally, these 1D latent tokens are decoded to reconstruct the frame $\hat{x}_t$. In addition, all three stages are guided by our context model, which integrates two types of context: long-term context from previous 1D latent tokens and short-term context from the preceding frame. 

\vspace{1mm}
\noindent \textbf{Encoding.}
Our method takes a video as input. For the current frame $x_t \in \mathbb{R} ^{3 \times H \times W}$,  it is first embedded into $Emb(x_t) \in \mathbb{R} ^{D \times h \times w}$ through patch embedding, with a patch size of $h = H/16$ and $w = W/16$. The embedded features $Emb(x_t)$ are then flattened into a 1D sequence $E_t \in \mathbb{R}^{D \times (h \cdot w)}$.
Next, The embedding $E_t$ is concatenated with learnable 1D latent tokens $L \in \mathbb{R}^{D \times (N \cdot 32)}$, where $N = \frac{h \times w}{16 \times 16}$, and the context $C$ from the context model, which integrates both long-term context $C_l \in \mathbb{R} ^{D \times (N \cdot 32)}$ and the flattened short-term context $C_s \in \mathbb{R} ^{D \times (h \cdot w)}$. The combined tokens are then fed into the encoder to generate compact 1D tokens $y_t$.

\vspace{-2mm}
\begin{gather}
    y_{t} = Enc(E_t \oplus L \oplus C), \\[0.25em]
    % C=\{C_l,C_s\}.
    C=C_l\oplus C_s.
\end{gather}

Specifically, our encoder consists of $M_e$ encoding blocks, and each block contains $N_e$ layers of local transformers followed by one layer of global transformer:

\vspace{-2mm}
\begin{equation}
Enc = \bigcup_{i=1}^{M_e} \{ \bigcup_{j=1}^{N_e}\text{LocalTrans}_{i,j}, \text{GlobalTrans}_{i} \}.
\end{equation}

For the local transformer, we partition the combined tokens into $N$ windows and process each window in parallel. Specifically, the embedding $E_t$ and short-term context $C_s$ are divided into $16 \times 16$ windows, while the 1D latent tokens $L$ and long-term context $C_l$ are divided into 32 tokens per window. Within each window, these four components are concatenated and fed into the local transformer, which leverages the Vision Transformer (ViT) to encode image tokens into 1D latent tokens. Through cascaded local transformer layers, the 1D latent tokens effectively aggregate key semantic and motion information from the image tokens and the context $C$, thereby producing a compact 1D token representation with rich semantics.

To model the correlations across windows, we introduce a global transformer layer after every $N_e$ local transformer layers. Specifically, the 2D image tokens $E_t$, short-term context $C_s$, 1D latent tokens $L$, and long-term context $C_l$ are first recombined and assigned 2D Rotary Positional Embeddings (RoPE)~\cite{heo2024rotary} to independently model relative positional relationships. The resulting tokens are then fed into the global transformer, which employs a ViT to process the entire sequence and facilitate global information exchange. This design allows independent windows to interact through global attention while maintaining low computational cost by employing only a few global transformer layers.

Consequently, we encode these compact 1D tokens to produce $y_t \in \mathbb{R} ^{D \times (N \cdot 32)}$. Unlike previous methods~\cite{qi2025generative} that encode frames into dense 2D latent grids, our approach compresses frames into 1D latent tokens. By removing rigid spatial correspondences, these tokens provide greater flexibility in modeling frames, capture richer semantics, and form a more compact latent space, thereby achieving lower bitrates and improved perceptual reconstruction quality.

\vspace{1mm}
\noindent \textbf{Entropy Model.}
To estimate the probability distribution of 1D latent tokens for arithmetic coding, we employ an autoregressive Transformer (AR Transformer)-based entropy model. Specifically, the 1D latent representation $y_t$ is first quantized into $Q(y_t)$, after which an AR Transformer models its distribution for entropy coding. During this process, the AR Transformer predicts the probability of each quantized token sequentially, conditioned on all previously decoded tokens. Consequently, the autoregressive modeling effectively captures inter-dependencies among tokens, thereby achieving  lower bitrates. Notably, the number of 1D latent tokens is small, and the 1D latent tokens from $N$ windows can be processed in parallel, keeping the computational cost of the AR Transformer manageable.

\vspace{1mm}
\noindent \textbf{Decoding.}
Our decoder follows an architecture similar to that of the encoder. Specifically, a sequence of learnable mask tokens $M \in \mathbb{R}^{D \times (h \cdot w)}$ is first integrated with the 1D latent tokens $\hat{y}_t$ and the context $C$. Through the decoding process, these mask tokens iteratively extract information from $\hat{y}_t$ and $C$, and the reconstructed frame $\hat{x}_t$ is then generated through a convolution output head.

\vspace{-2mm}
\begin{equation}
    \hat{x}_t = Out(Dec(\hat{y}_t \oplus M \oplus C)).
\end{equation} 

\begin{figure}[t]
% \vspace{-2mm}
    \centering
    \includegraphics[width=1\linewidth]{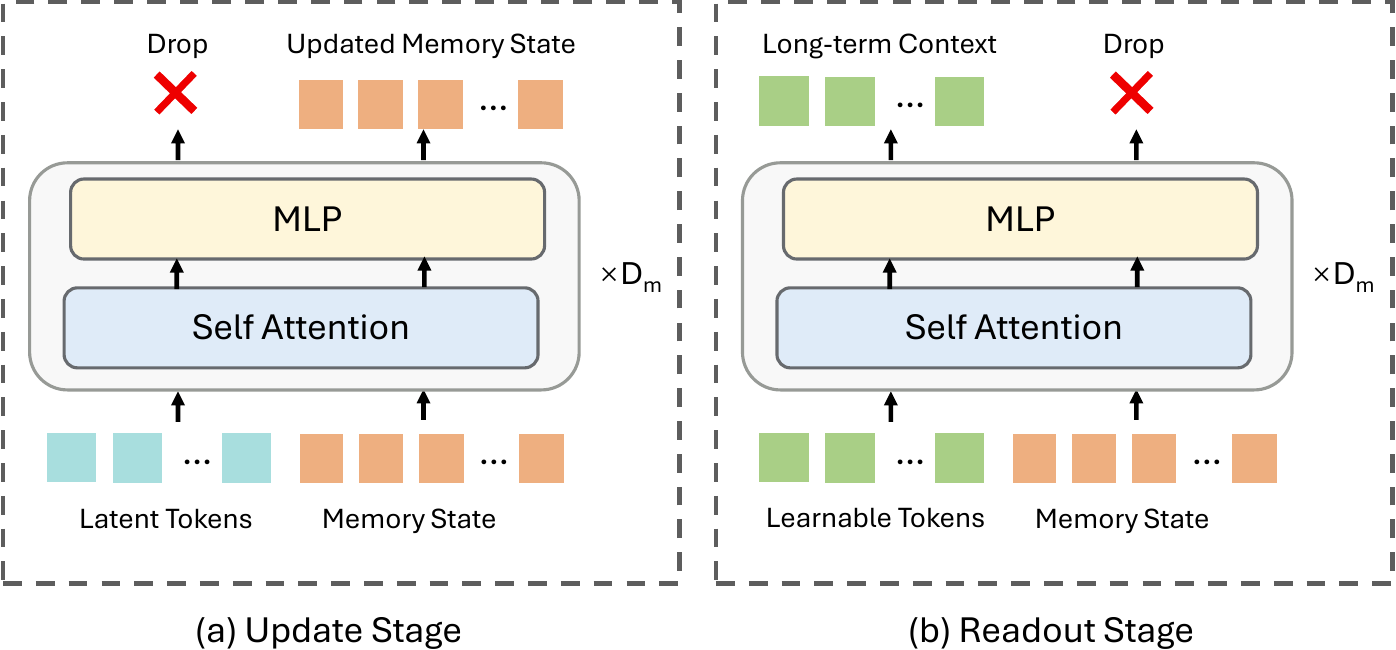}
      \vspace{-4mm}
    \caption{1D memory. $D_m$ denote the number of Transformer layers. We employ simple yet effective Transformer layers to manage long-term context.}
    \label{fig:memory}
     \vspace{-4mm}
\end{figure}

\subsection{Context Model}
\label{sec:methdo- memory}

In this section, we present our context model, which consists of two components: a long-term 1D memory and a short-term context buffer. Unlike previous methods~\cite{li2024neural, qi2025generative, jia2025towards}, our model incorporates long-term context derived from 1D tokens to provide richer temporal information. The short-term context captures fine grained features but lacks awareness of long-range dependencies, whereas the long-term context delivers global semantic context that supplements the short-term context. Together, these two components form a comprehensive temporal representation, effectively exploiting temporal dependencies and further reducing redundancy in videos.

\begin{figure}[t]
% \vspace{-2mm}
    \centering
    \includegraphics[width=1\linewidth]{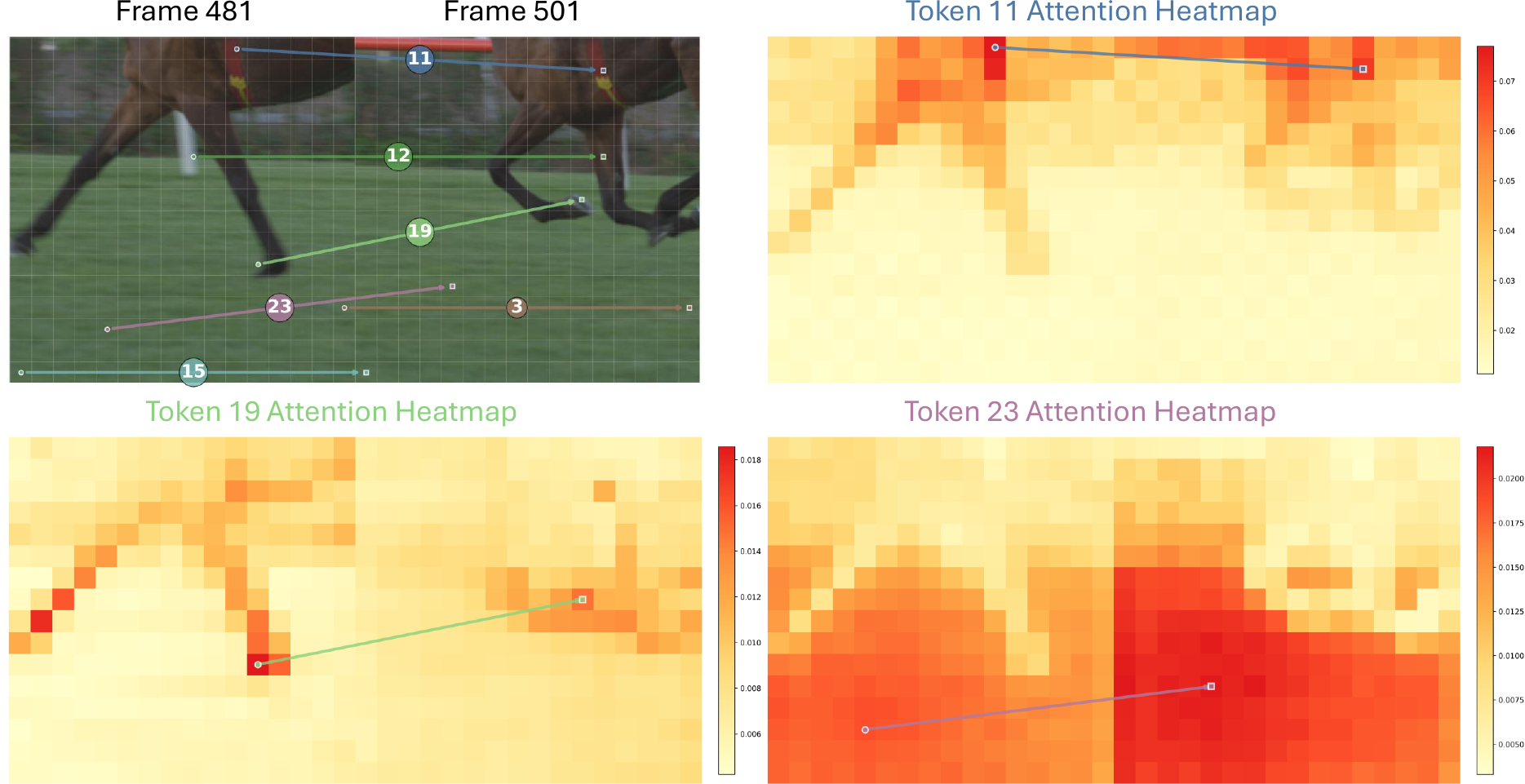}
      \vspace{-4mm}
    \caption{Visualization of 1D latent token outflows across two frames during object motion. In the two figures, the lines connect points corresponding to the maximum attention weights of each token, with the numbers indicating token indices (e.g., token 19 focuses on the horse’s left foreleg in both frames). Detailed attention maps show that each token consistently focuses on the same semantic region across frames, effectively capturing object motion.}
    \label{fig:object_motion_flow}
     \vspace{-4mm}
\end{figure}

\begin{figure}[t]
% \vspace{-2mm}
    \centering
    \includegraphics[width=1\linewidth]{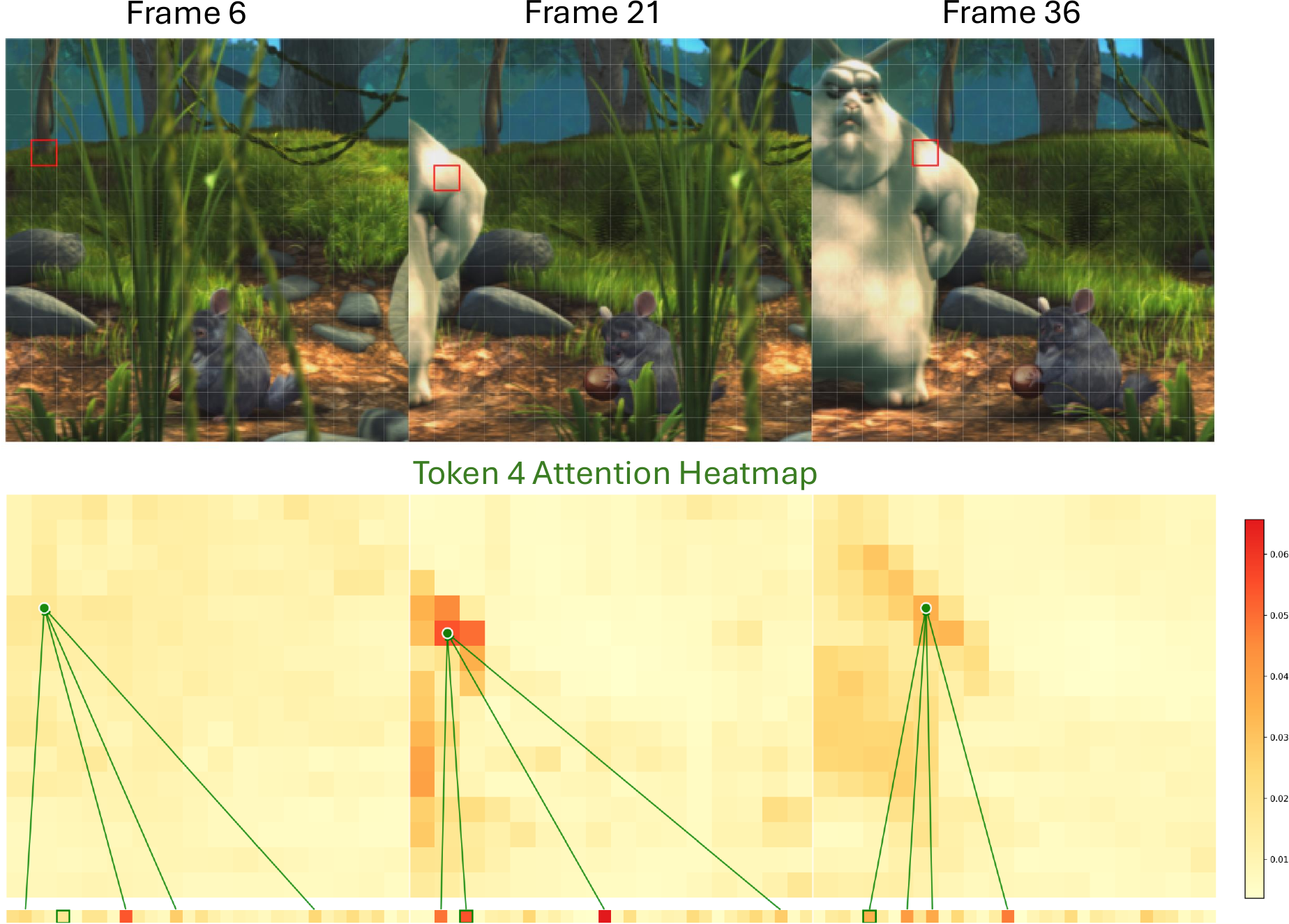}
      \vspace{-4mm}
    \caption{Visualization of the outflow variation of a 1D latent token (index 4) as a new object appears. The \textcolor{red}{red boxes} in the first row mark image patches with the highest attention weights, while the \textcolor{green}{green lines} in the second row link them to the top four 1D latent tokens with the strongest attention. The bottom row is the 1D latent tokens attention weights corresponding to the maximum weight image patch (red boxes). As new content emerges, attention weights gradually shift from previously active tokens to newly activated ones.}
    \label{fig:new_content_flow}
     \vspace{-4mm}
\end{figure}

\begin{figure*}[t]
    \begin{center}
        \includegraphics[width=1.0\linewidth]{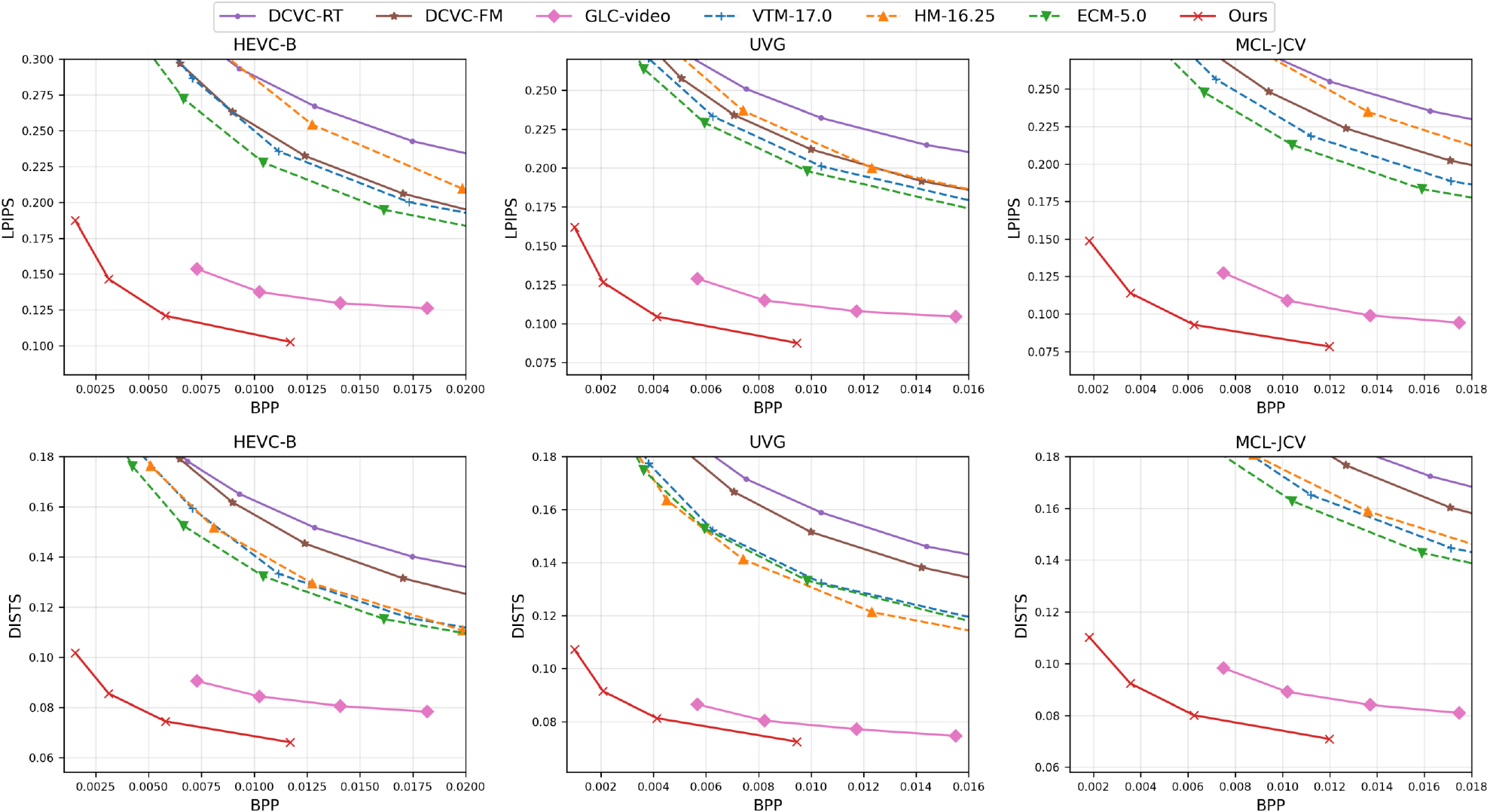}
    \end{center}
    \vspace{-5mm}
    \caption{
        Rate-distortion curves on the HEVC-B, the UVG and the MCL-JCV datasets. 96 frames are tested in RGB colorspace with intra-period=–1.
    }
    \label{fig:exp1024}
    % \vspace{-2mm}
\end{figure*}

For the short-term context, similar to previous methods~\cite{li2024neural}, we use the decoder features $f_t$ from the preceding frame to provide detailed short-term context. 
For the long-term context, we design a simple yet effective 1D memory that provides global semantic features using the Transformer architecture. 
As shown in Fig.~\ref{fig:memory}, our 1D memory maintains a fixed size memory state and comprises two stages for recurrently updating the memory and reading out long-term context. In the update stage, a few 1D latent tokens $\hat{y}_t$ are used to update the memory state, while in the readout stage, learnable query tokens retrieve the long-term context from the memory. All operations are performed in the 1D latent space, which is free from rigid 2D spatial constraints and thus enables flexible memory management without complex designs. By leveraging compact and semantically rich 1D latent tokens to manage memory, our 1D memory provides more global and semantic context than the fine-grained short-term context. Moreover, since the number of 1D latent tokens is considerably smaller than that of the 2D latent grids used in previous works, the model can store more information within a fixed memory capacity, effectively alleviating information forgetting.

\subsection{Analysis on 1D Latent Tokens}
\label{sec:methdo-Attention map visualization}

To validate that our 1D representation can effectively and flexibly model videos in a semantic manner, thereby achieving better perceptual quality, we follow FlowCut~\cite{tong2025flowcut} and visualize the attention maps of the 1D latent tokens corresponding to the final reconstructed frames. Specifically, for an attention map $ A \in \mathbb{R}^{M \times M}$, we visualize the outflow of each 1D latent token, where the outflow represents the primary destination of its information, defined as $argmax A[:, j]$. We illustrate two two representative video scenarios: object motion and the appearance of new content, based on the attention map from the last decoder layer.

\vspace{1mm}
\noindent  \textbf{Object Motion.}
As shown in Fig.~\ref{fig:object_motion_flow}, each token consistently focuses on the same semantic region across two frames, even under large motion. For example, token 23 attends to the grassland in both frames. In more challenging cases involving large motion, such as token 19 focusing on the horse’s left foreleg in both frames, more detailed information is required. Consequently, the attention map of token 19 covers a smaller area than that of token 23. 
These observations demonstrate that our 1D latent token design effectively tracks consistent semantic regions throughout the video while capturing intrinsic motion. Moreover, it can adaptively assign attention weights across the entire image without being constrained by the 2D spatial structure.

\vspace{1mm}
\noindent  \textbf{New Content.}
As shown in Fig.~\ref{fig:new_content_flow}, we visualize the attention maps of token 4 and the variation in attention weights of the 1D latent tokens corresponding to newly appeared content. Specifically, as the rabbit gradually emerges in the video, the attention map shifts its focus from the blank background to the rabbit. 
Initially, the token focuses on the grass background. As the rabbit emerges, the attention weights of certain 1D latent tokens progressively increase to capture the new content, indicating that they begin to attend to the newly appeared object. Finally, the attention weights are re-assigned to adapt to the new scene. This observation demonstrates the effectiveness of our 1D latent token design in handling the emergence of new objects, as the 1D tokens can dynamically reallocate attention to newly appearing content.
\section{Experiment}
\label{sec:exp}

\begin{figure*}[t]
    \begin{center}
        \includegraphics[width=1.0\linewidth]{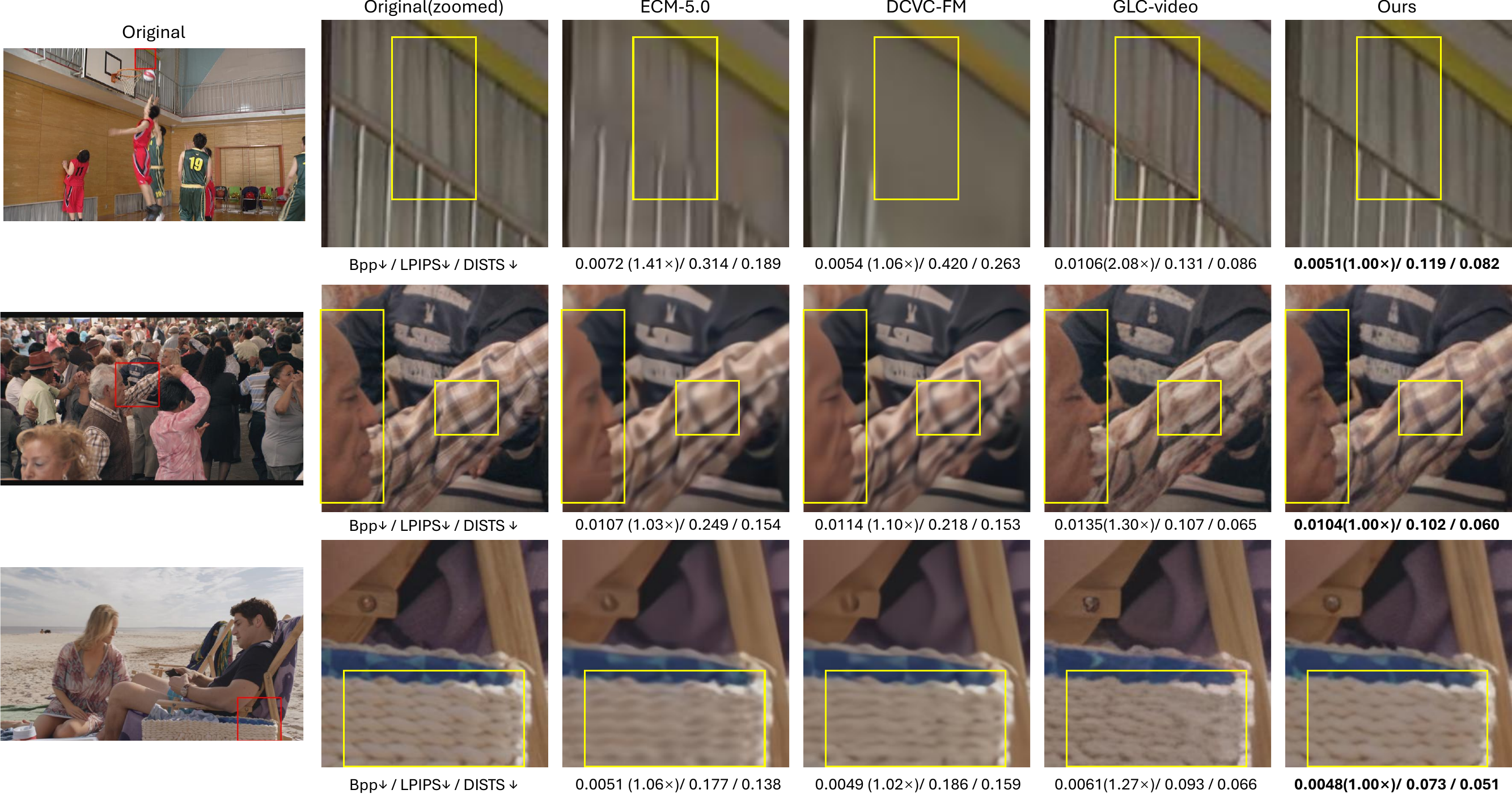}
    \end{center}
    \vspace{-2mm}
    \caption{
        Qualitative examples on the MCL-JCV datasets.
        % Qualitative examples on the MCL-JCV dataset.
    }
    \label{fig:visual1024}
    % \vspace{-3mm}
\end{figure*}

\subsection{Experimental Protocol}
\label{sec:experiments-settings}

\noindent \textbf{Training Details.}
Our model is trained on the Vimeo~\cite{vimeo,ori_vimeo} and OpenVid-HD~\cite{nan2025openvidm} datasets.
We train a variable-bitrate model using the loss function $L=R+\lambda D$, where $R$ and $D$ denote the bitate and distion, respectively. The trade-off parameter $\lambda$ is sampled at 8 points using log-space linear interpolation within the range $[0.07,1.5]$. 
The detailed training strategy and loss formulation are provided in the supplementary material.

\vspace{1mm}
\noindent \textbf{Test Details.}
We test our model on the HEVC-B~\cite{flynn16common}, UVG~\cite{uvg}, and MCL-JCV~\cite{mcl-jcv} datasets. 
All experiments are conducted under low-delay encoding settings(intra-period=–1) in the RGB colorspace. The BD-Rate~\cite{bd-rate} metric is employed for performance evaluation.
To assess reconstruction quality, we use the reference perceptual metrics LPIPS~\cite{zhang2018unreasonable} and DISTS~\cite{ding2020image}. The bitrate is measured in bits per pixel (bpp), and the PSNR and MS-SSIM~\cite{wang2003multiscale} results are also reported in the supplementary material.

\vspace{1mm}
\noindent \textbf{Baselines.}
For traditional codecs, we compare against HM-16.25~\cite{HM}, VTM-17.0~\cite{VTM}, and ECM-5.0~\cite{ECM}, which are primarily optimized for PSNR. For neural methods, we include the PSNR-oriented codecs DCVC-FM~\cite{li2024neural} and DCVC-RT~\cite{jia2025towards}, as well as the perceptual generative codec GLC-Video~\cite{qi2025generative}, which targets perceptual quality..

\subsection{Experimental Results}
\label{sec:Main Results}
\noindent \textbf{Quantitative Evaluation.}
% In Fig.~\ref{fig:exp1024}, we present the rate-distortion curves across multiple datasets. For perceptual metrics, our method consistently outperforms all other methods, achieving substantially lower LPIPS and DISTS at the same bitrates.
% Table~\ref{tab:1024} reports the BD-Rate results on the LPIPS and DISTS metrics across multiple datasets. Compared with the perceptual video codec GLC-Video, our method achieves bitrate savings of 60.4\% for LPIPS and 68.8\% for DISTS on the HEVC-B dataset, demonstrating the potential of our 1D latent token design in perceptual video compression. By adopting 1D latent tokens that eliminate the 2D structure, reducing the token count and introducing 1D-based long-term context, our model achieves lower bitrates while maintaining high perceptual quality.
In Fig.~\ref{fig:exp1024}, we present the rate-distortion curves across multiple datasets. For perceptual metrics, our method consistently outperforms all other methods, achieving substantially lower LPIPS and DISTS at the same bitrates.
Table~\ref{tab:1024} reports the BD-Rate results across multiple datasets.
For perceptual metrics (LPIPS and DISTS), compared with GLC-Video, our method achieves bitrate savings of 60.4\% for LPIPS and 68.8\% for DISTS on the HEVC-B dataset.
For fidelity metrics (PSNR and MS-SSIM), our method achieves bitrate savings of 53.8\% for PSNR and 45.1\% for MS-SSIM on the HEVC-B dataset compared with GLC-Video. Detailed rate-distortion curves for PSNR and MS-SSIM are provided in the supplementary material.
These results demonstrate the potential of our 1D latent token design in perceptual video compression. By eliminating the 2D structure, reducing the token count, and introducing 1D-based long-term context, our model achieves lower bitrates while maintaining high perceptual quality.
\begin{table}[t]
\begin{center}
\caption{
    BD-Rate (\%) comparison in the RGB colorspace on the HEVC-B, UVG, and MCL-JCV datasets (lower is better). 96 frames are evaluated with an intra-period of –1.
}
% \vspace{-2mm}
\label{tab:1024}
\resizebox{0.49\textwidth}{!}{
\begin{tabular}{l|cccccc}
\hline
\multirow{2}{*}{Method}             & \multicolumn{2}{c}{HEVC-B}            & \multicolumn{2}{c}{UVG}            & \multicolumn{2}{c}{MCL-JCV}    \\
                                    & LPIPS            & DISTS              & LPIPS            & DISTS           & LPIPS            & DISTS       \\ \hline
GLC-video~\cite{qi2025generative}   & 0.0\%            & 0.0\%              & 0.0\%            & 0.0\%           & 0.0\%            & 0.0\%           \\ 
Ours                                & -60.4\%          & -68.8\%            & -66.0\%          & -47.9\%         & -62.1\%          & -61.5\%         \\ 
\hline
\end{tabular}
}

\vspace{3mm}

\resizebox{0.49\textwidth}{!}{
\begin{tabular}{l|cccccc}
\hline
\multirow{2}{*}{Method}             & \multicolumn{2}{c}{HEVC-B}            & \multicolumn{2}{c}{UVG}            & \multicolumn{2}{c}{MCL-JCV}    \\
                                    & PSNR            & MS-SSIM              & PSNR            & MS-SSIM           & PSNR            & MS-SSIM       \\ \hline
GLC-video~\cite{qi2025generative}   & 0.0\%            & 0.0\%              & 0.0\%            & 0.0\%           & 0.0\%            & 0.0\%           \\ 
Ours                                & -53.8\%          & -45.1\%            & -62.5\%          & -45.8\%         & -54.7\%          & -33.0\%         \\ 
\hline
\end{tabular}
}
\end{center}
\vspace{-2mm}
\end{table}

\vspace{1mm}
\noindent  \textbf{Qualitative Evaluation.}
In Fig.~\ref{fig:visual1024}, we present visual comparisons on the UVG and MCL-JCV datasets. The neural video codec DCVC-FM and the traditional codec ECM-5.0 struggle to reproduce realistic details at low bitrates, leading to blurred regions (e.g., the railing in the first example and the basket pattern in the third example). The perceptual codec GLC-Video enhances visual realism but introduces noticeable artifacts and distortions, such as unrealistic facial features and clothing textures in the second example. In contrast, our method produces reconstructions with fewer artifacts, more accurate colors, and finer details, even at lower bitrates.

\begin{table}[t]
\begin{center}
\caption{
    BD-Rate comparison for different token sizes. The token size of $32 \times 16$ is used as the anchor.
}
\vspace{-2mm}
\label{tab:ablation_tokens}
\resizebox{0.40\textwidth}{!}{
\begin{tabular}{l|ccc}
\hline
Token Size        & \multirow{2}{*}{HEVC-B}            & \multirow{2}{*}{UVG}      & \multirow{2}{*}{MCL-JCV}    \\
(Number $\times$ Channel)        \\ \hline
$16 \times 32$                    & 41.7\%       & 47.6\%       & 37.3\%      \\
$16 \times 64$                    & 20.1\%      & 23.4\%       & 12.2\%     \\
$32 \times 16$(\textbf{Ours})     & 0.0\%        & 0.0\%        & 0.0\%       \\
$32 \times 32$                    & 14.2\%       & 21.5\%     & 10.2\%     \\
$64 \times 16$                    & 24.2\%      & 22.8\%       & 17.6\%      \\ \hline

\end{tabular}
}
\end{center}
\vspace{-2mm}
\end{table}

\begin{table}[t]
\begin{center}
\caption{
    BD-Rate comparison of different model variants. AR denotes the autoregressive entropy model, and Memory denotes the memory component, where 1D and 2D indicate the use of 1D or 2D features to manage the memory. We use setting (4) as the anchor, which is finally adopted by our method.
}
\vspace{-2mm}
\label{tab:ablation_arch}
\resizebox{0.40\textwidth}{!}{
\begin{tabular}{ccc|ccc}
\hline
Setting   & AR  & Memory    & HEVC-B          & UVG     & MCL-JCV    \\ \hline
(1)      & \ding{55}    & \ding{55}            & 67.8\%        & 67.4\%       & 41.8\%     \\ 
(2)      & \ding{51}    & \ding{55}            & 20.1\%        & 40.6\%       & 24.2\%    \\ 
(3)      & \ding{51}    & 2D            & 11.5\%        & 16.8\%       & 7.3\%    \\ 
(4)      & \ding{51}    & 1D            & 0.0\%         & 0.0\%     & 0.0\%   \\  \hline

\end{tabular}
}
\end{center}
\vspace{-2mm}
\end{table}

\subsection{Ablation Study}
\label{sec:experiments-ablation}
In this section, we conduct ablation studies on different latent token sizes, the 1D memory, and the autoregressive entropy model. All model variants are evaluated on the HEVC-B, UVG, and MCL-JCV datasets, as measured by LPIPS metric.

\vspace{1mm}
\noindent  \textbf{1D Latent Token Size.}
We evaluate the impact of different 1D latent token sizes. As shown in Table~\ref{tab:ablation_tokens}, a smaller token size (e.g., $16 \times 32$, where 16 denotes the number of tokens and 32 denotes the token channel) restricts the model’s capacity, thereby degrading reconstruction quality and resulting in a worse BD-Rate. In contrast, a larger token size (e.g., $64 \times 16$) increases the bitrate due to its higher dimensionality, which also leads to a worse BD-Rate. Hence, selecting an appropriate token size is essential to achieve a balanced trade-off between bitrate and reconstruction quality.

\vspace{1mm}
\noindent  \textbf{1D Memory.}
In this variant, we first remove the 1D memory to assess the importance of long-term contextual information. As shown in Table~\ref{tab:ablation_arch}, comparing settings (2) and (4) reveals that removing the long-term context results in a BD-Rate increase of over 40\% on the UVG dataset, demonstrating the necessity of long-term context. Next, we replace the 1D memory with a 2D memory constructed from 2D short-term context to examine the effectiveness of the 1D memory design. By comparing settings (3) and (4), we observe that this replacement leads to a BD-Rate degradation of more than 16\% on the UVG dataset, confirming that using 1D features for memory management is more effective.

\vspace{1mm}
\noindent  \textbf{Autoregressive Entropy Model.}
We further evaluate the effectiveness of our proposed autoregressive entropy model by replacing it with a non-autoregressive version, under the same setting without the 1D memory. As shown in Table~\ref{tab:ablation_arch}, comparing setting (1) and (2), replacing the autoregressive entropy model leads to a worse BD-Rate. These results confirm that the autoregressive entropy model effectively reduces redundancy among tokens.

\subsection{Complexity Analysis}
\label{sec:experiments-complexity}

\begin{table}[t]
\begin{center}
\caption{
    Complexity analysis using fp16 precision at 1080P resolution.
    The tests are conducted on an NVIDIA A100 GPU.
}
\vspace{-2mm}
\label{table:complexity}
\resizebox{0.35\textwidth}{!}{
\begin{tabular}{l|cc}
\hline
Model                                     & Enc. Time (s)          & Dec. Time (s)       \\ \hline
DCVC-FM~\cite{li2024neural}               & 0.183                  & 0.190         \\
GLC-video~\cite{qi2025generative}         & 0.290                  & 0.183           \\
Ours                                      & 0.262                  & 0.207          \\ \hline
\end{tabular}
}
\end{center}
\vspace{-5mm}
\end{table}

We compare the average coding time in Table~\ref{table:complexity}.
Compared with DCVC-FM, our method is only slightly slower due to its larger model size. However, the larger model provides stronger capacity, enabling our method to achieve significantly better perceptual quality with richer details.
Compared with GLC-Video, our method attains a similar overall coding speed, with the encoding even slightly faster, while delivering notably higher visual quality.
Overall, our approach improves compression efficiency while maintaining a relatively fast coding speed. This efficiency stems from performing compression in a compact latent space with fewer tokens, utilizing a parallel window processing design, and employing advanced techniques such as FlashAttention~\cite{dao2023flashattention2} to accelerate computation.

\section{Conclusion} 
\label{sec:conclusion}
In this paper, we present a novel method for modeling videos using 1D latent tokens instead of conventional dense 2D latent grids. We first analyze the limitations of 2D latent grids, including the rigid 2D spatial structure and the redundancy of token numbers. To address these issues, we represent videos with 1D latent tokens. Without the constraints of the rigid spatial correspondences, 1D latent tokens can effectively capture intrinsic object motion and semantics while flexibly reducing the number of tokens. Moreover, our 1D memory incorporates semantical long-term context to further exploit temporal information. By leveraging compact and semantically rich 1D latent tokens to manage memory, our design effectively mitigates information forgetting while maintaining low computational cost. Experimental results demonstrate that our model achieves state-of-the-art reconstruction quality while maintaining a relatively fast coding speed.

\vspace{1mm}
\noindent  \textbf{Limitations.}
Due to the inherent information loss from representing each frame with only 32 1D tokens and the limited token capacity, our model currently targets lossy compression at low bitrates. Future work will explore extending it towards lossless video compression.
{
    \small
    \bibliographystyle{ieeenat_fullname}
    \bibliography{main}
}
\clearpage
\maketitlesupplementary
\appendix
\section{Training Details}
This section provides the training details of our model. We train the model on the Vimeo~\cite{vimeo,ori_vimeo} and OpenVid-HD~\cite{nan2025openvidm} datasets. The OpenVid-HD dataset contains 433,523 videos. For the Vimeo dataset, in addition to the 7-frame sequences in Vimeo-90k, we follow DCVC-FM~\cite{li2024neural} to process the raw Vimeo videos and generate approximately 6,000 additional 32-frame sequences. Specifically, we use the original Vimeo videos~\cite{ori_vimeo}, apply scene detection, and randomly sample 32-frame clips.
Our training strategy is as follows. The model is optimized using the AdamW optimizer~\cite{loshchilov2017decoupled} throughout the entire training process.

\vspace{1mm}
\noindent \textbf{The first stage.} We train only the encoder and decoder using two-frame samples from the OpenVid-HD dataset, resized to $256\times256$, with a batch size of 128. We adopt the reconstruction loss function from TA-Titok~\cite{kim2025democratizing}:
\begin{equation}
\begin{aligned}
L_{\text{stage1}} = & \|x-\hat{x}\|_{2} 
   + \\
   & \lambda_{\text{LPIPS}} L_{\text{LPIPS}}(x, \hat{x}) + \\
   & \lambda_{\text{adv}} L_{\text{adv}}(x, \hat{x}).
\end{aligned}
\end{equation}
Here, $L_{\text{LPIPS}}$ denotes the perceptual loss, and $L_{\text{adv}}$ denotes the adversarial loss~\cite{goodfellow2020generative}. The corresponding weighting coefficients, $\lambda_{\text{LPIPS}}$ and $\lambda_{\text{adv}}$, are set to $1.1$ and $0.1$, respectively, following TA-Titok. Consistent with TA-Titok, we use a maximum learning rate of $1\mathrm{e}{-4}$ and apply a cosine learning rate schedule.
Both the encoder and decoder are initialized with TA-Titok pretrained weights, which provide a strong prior and facilitate faster convergence.

\noindent \textbf{The second stage.} We introduce the entropy model and train the network on resized $256\times256$ videos from the Vimeo dataset. Regarding the AR entropy model, we adopt a standard Transformer for token-by-token probability prediction. As for the non-autoregressive version used for ablation experiments, we adopt the two-step entropy model from DCVC-RT~\cite{jia2025towards}.
The number of training frames progressively increases from 2 to 32, while the batch size decreases from 128 to 4 according to the training frames. The learning rate is set to $5\mathrm{e}{-5}$. The loss function is defined as:
\begin{equation}
    L_{\text{stage2}} =\frac{1}{T}\sum_{t=1}^{t=T}{(R + \lambda L_{\text{stage1}})},
\end{equation}
where $R$ denotes the bitrate and $T$ denotes the number of frames used in cascade training. The trade-off parameter $\lambda$ is sampled at eight points using logarithmic-space linear interpolation within the range $[0.07, 1.5]$. 

\noindent \textbf{The third stage.} We sequentially introduce the global Transformer and 1D memory, and finetune the model on OpenVid-HD videos with variable resolutions ranging from $256 \times 256$ to $1280 \times 2048$. The training sequence length is set to 32 frames with a batch size of 2. The learning rate is set to $1\mathrm{e}{-5}$. To reduce GPU memory consumption, following ECVC~\cite{jiang2025ecvc}, we employ the partial cascaded finetuning strategy with a frame group size of 4. The loss function remains the same as in the second stage.

\section{Experiments}
This section describes the evaluation settings and presents additional quantitative and qualitative results.

\subsection{Evaluation Details}
We evaluate all models on the 96-frame videos from the HEVC-B~\cite{flynn16common}, UVG~\cite{uvg}, and MCL-JCV~\cite{mcl-jcv} datasets. All videos are processed so that both their height and width are multiples of 256.

\begin{figure*}[t]
    \centering
    \includegraphics[width=1.0\linewidth]{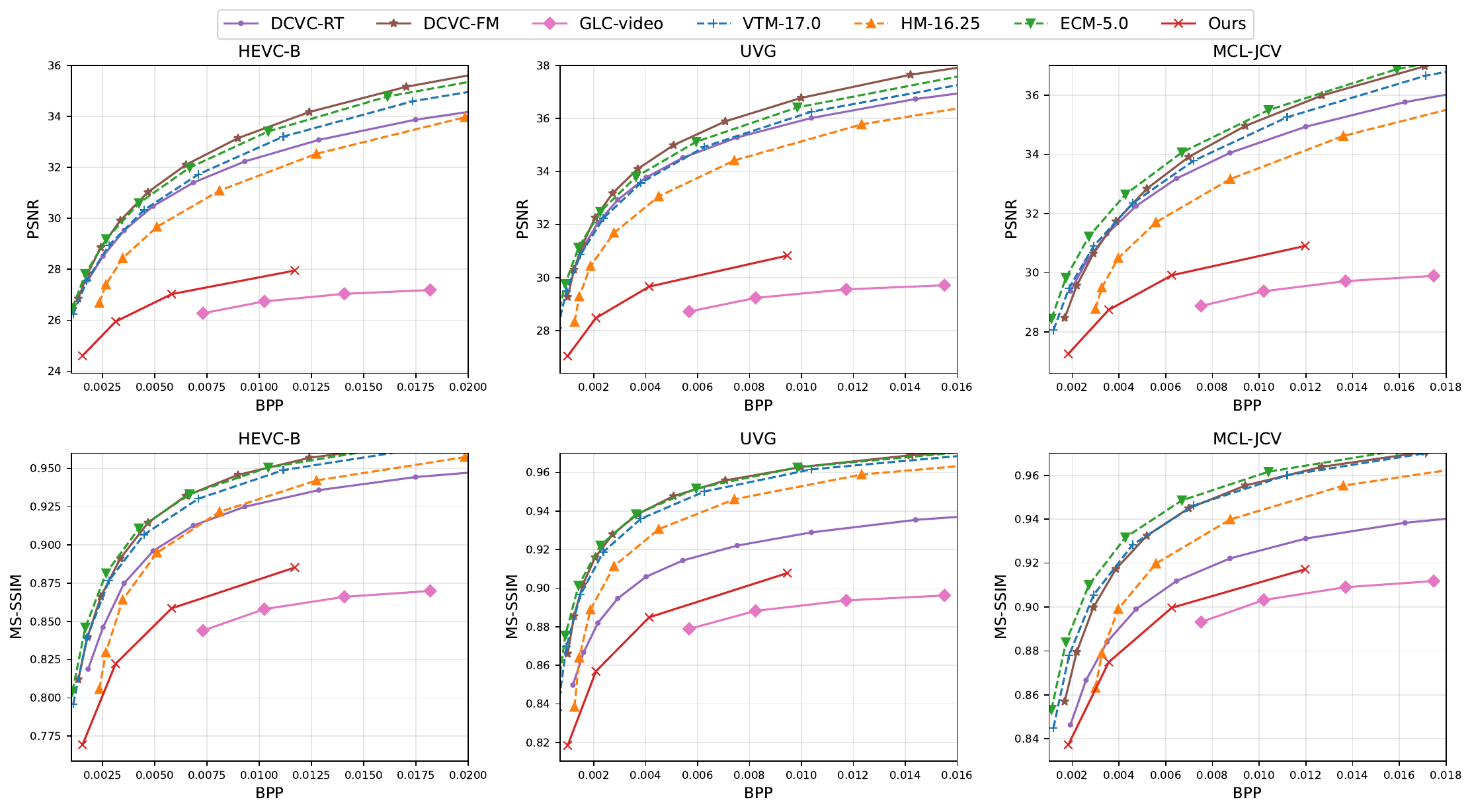}
    \caption{Rate-distortion curves in terms of PSNR and MS-SSIM.
}
    \label{fig:psnr_curve}
\end{figure*}

\begin{figure*}[t]
    \centering
    \includegraphics[width=1.0\linewidth]{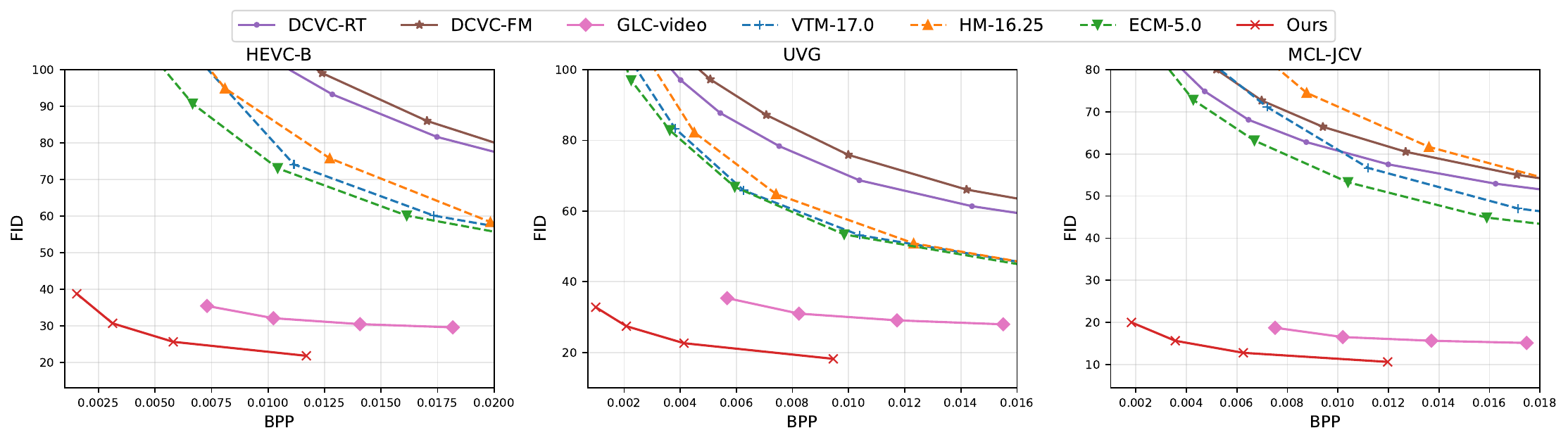}
    \caption{Rate-distortion curves in terms of FID.
}
    \label{fig:fid_curve}
\end{figure*}

For traditional video codecs, we evaluate VTM-17.0~\cite{VTM}, HM-16.25~\cite{HM}, and ECM-5.0~\cite{ECM} in the RGB colorspace, following the protocol used in the DCVC-FM~\cite{li2024neural}, which converts the RGB input to 10-bit YUV444 for internal codec processing and adopts the low-delay encoding setting. We use the official configuration files for VTM, HM, and ECM, respectively: \textit{encoder\_lowdelay\_vtm.cfg}, \textit{encoder\_lowdelay\_main\_rext.cfg}, and \textit{encoder\_lowdelay\_ecm.cfg}.
The parameters for encoding are as:
\begin{myitemize}
    \item
	-c \{{\em config file name}\}\par
	-\/-InputFile=\{{\em input file name}\}\par
	-\/-InputBitDepth=10\par
	-\/-OutputBitDepth=10 \par
	-\/-OutputBitDepthC=10 \par
	-\/-InputChromaFormat=444\par
	-\/-FrameRate=\{{\em frame rate}\}\par
	-\/-DecodingRefreshType=2\par
	-\/-FramesToBeEncoded=\{{\em frame number}\}\par
	-\/-SourceWidth=\{{\em width}\}\par
	-\/-SourceHeight=\{{\em height}\}\par
	-\/-IntraPeriod=\{{\em intra period}\}\par
	-\/-QP=\{{\em qp}\}\par
	-\/-Level=6.2\par
	-\/-BitstreamFile=\{{\em bitstream file name}\}\par
\end{myitemize}

For previous neural video codecs~\cite{li2024neural,jia2025towards,qi2025generative}, and our method, we adopt low-delay encoding settings (intra-period = –1) in the RGB colorspace.

\begin{figure*}[t]
    \centering
    \includegraphics[width=1.0\linewidth]{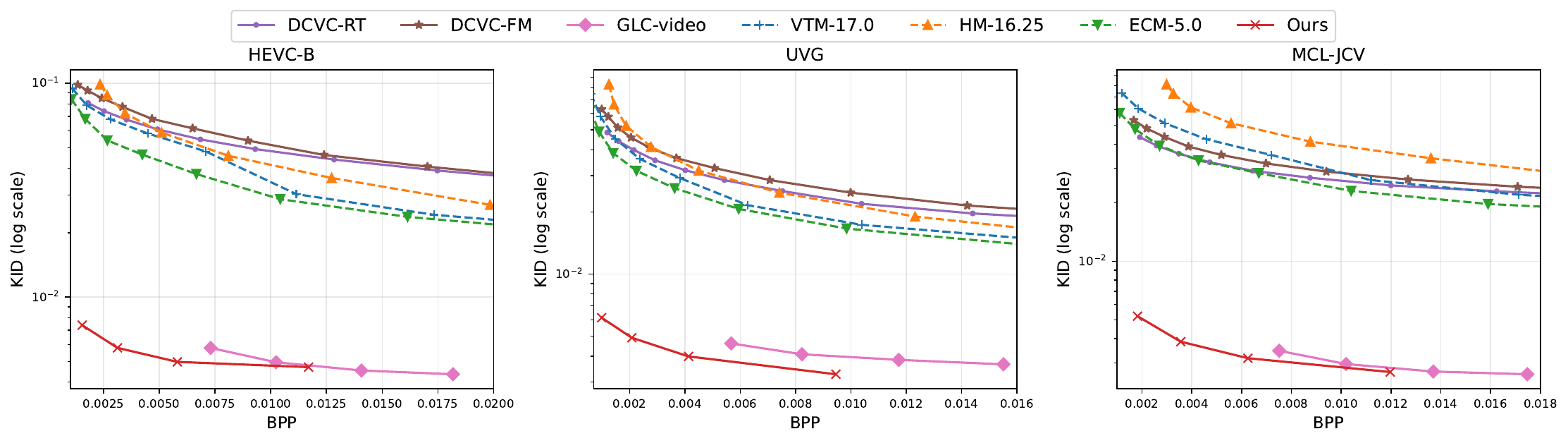}
    \caption{Rate-distortion curves in terms of KID.
}
    \label{fig:kid_curve}
\end{figure*}

\begin{figure*}[t]
    \centering
    \includegraphics[width=1.0\linewidth]{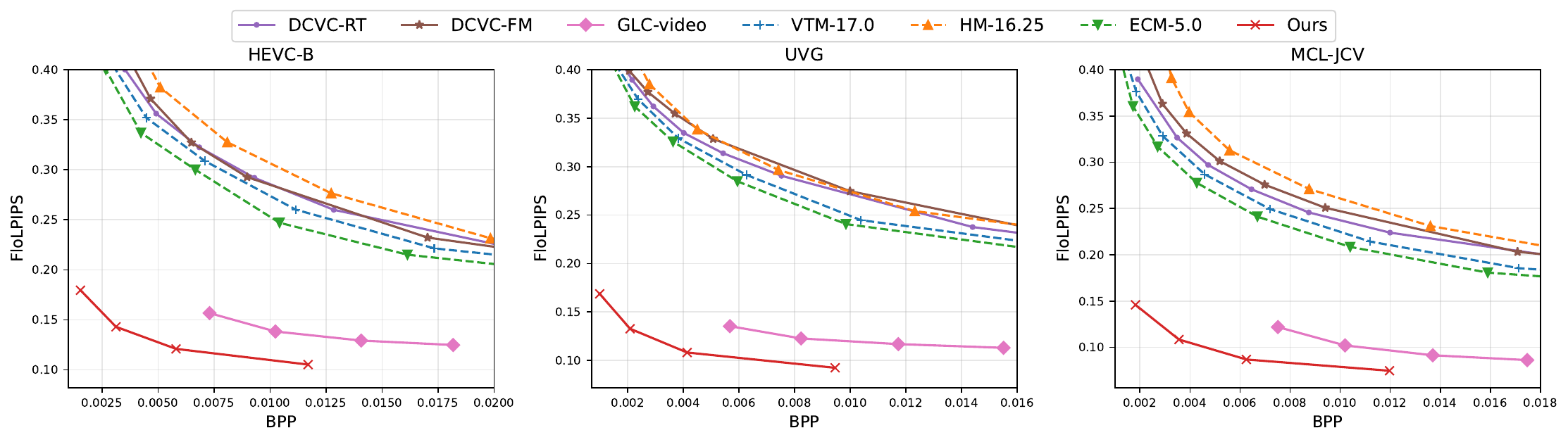}
    \caption{Rate-distortion curves in terms of FloLPIPS.
}
    \label{fig:flolpips_curve}
\end{figure*}

\subsection{Quantitative Results}
We report traditional pixel-level distortion metrics, PSNR and MS-SSIM~\cite{wang2003multiscale} results in Fig.~\ref{fig:psnr_curve}. 
Compared with GLC-video~\cite{qi2025generative}, which is similarly optimized for perceptual quality, our method not only achieves better perceptual metrics (LPIPS and DISTS) but also attains higher objective metrics (PSNR and MS-SSIM). These results confirm the effectiveness of our approach for perceptual video compression.
Although our method produces lower PSNR than objective video codecs~\cite{li2024neural,ECM}, both the visual comparisons and the perceptual metrics presented in the main text consistently demonstrate its superior perceptual performance. Moreover, at such low bitrates, objective codecs (typically optimized for PSNR) tend to lose fine details, resulting in blurry reconstructions. Therefore, higher PSNR or MS-SSIM at these bitrates does not necessarily correspond to better perceptual quality.

We further report the FID~\cite{heusel2017gans} metric in Fig.~\ref{fig:fid_curve} and KID~\cite{sutherland2018demystifying} metric in Fig.~\ref{fig:kid_curve}. The FID and KID metrics are evaluated by splitting the frames into overlapped $256 \times 256$ patches, following the method in~\cite{xue2025dlf}. FID and KID measure the distribution-level similarity between reconstructed and ground-truth frames, thus providing complementary insights beyond pixel- or feature-level metrics. As shown in the figures, our method consistently outperforms all competing approaches across both metrics, indicating that our reconstructions better align with the statistical characteristics of natural video distributions and exhibit superior overall fidelity.

\begin{figure*}[t]
    \centering
    \includegraphics[width=1.0\linewidth]{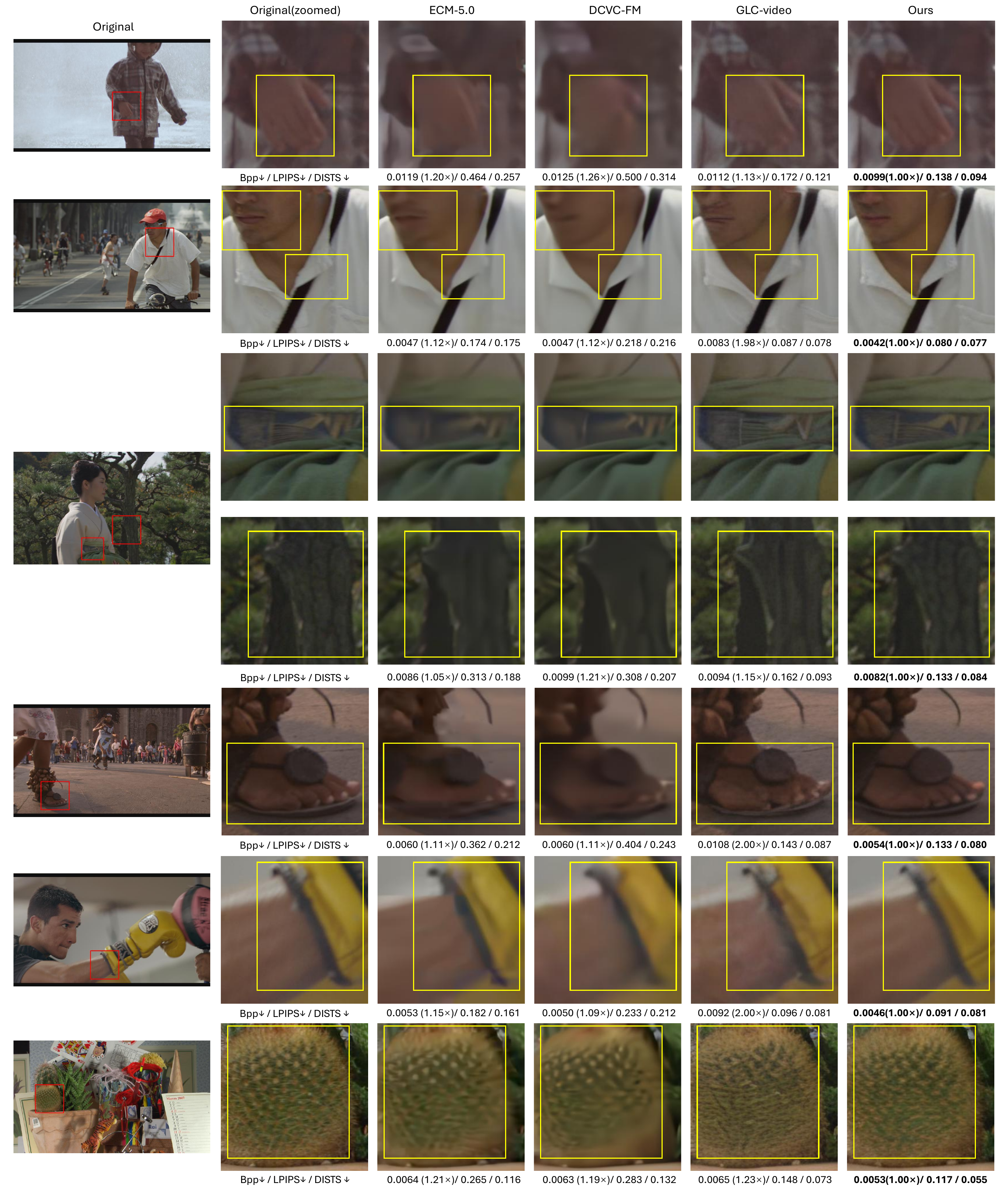}
    \caption{More visual examples.}
    \label{fig:suppl_visual}
\end{figure*}

\subsection{Qualitative Results}
In this section, we present additional qualitative results from the HEVC-B, UVG, and MCL-JCV datasets (Fig.~\ref{fig:suppl_visual}). Our method produces visually sharper and more realistic reconstructions compared with other baselines. These examples further demonstrate that our approach excels not only in quantitative metrics but also in preserving fine structures, color consistency, and perceptual naturalness, thereby delivering superior visual quality.

\section{Temporal Consistency Evaluation}
In this section, we test the temporal consistency of our method through both quantitative and qualitative evaluation.

For quantitative results, we report the FloLPIPS~\cite{danier2022flolpips} metric in Fig.~\ref{fig:flolpips_curve}. FloLPIPS is a flow-guided variant of LPIPS designed to measure temporal perceptual consistency in videos. As shown in the figure, our method consistently achieves lower FloLPIPS scores across all datasets, indicating better temporal coherence and reduced flickering artifacts. This further validates that the proposed method not only enhances perceptual quality but also preserves inter-frame consistency, which is crucial for producing visually stable video results.

\begin{figure*}[t]
    \centering
    \includegraphics[width=0.8\linewidth]{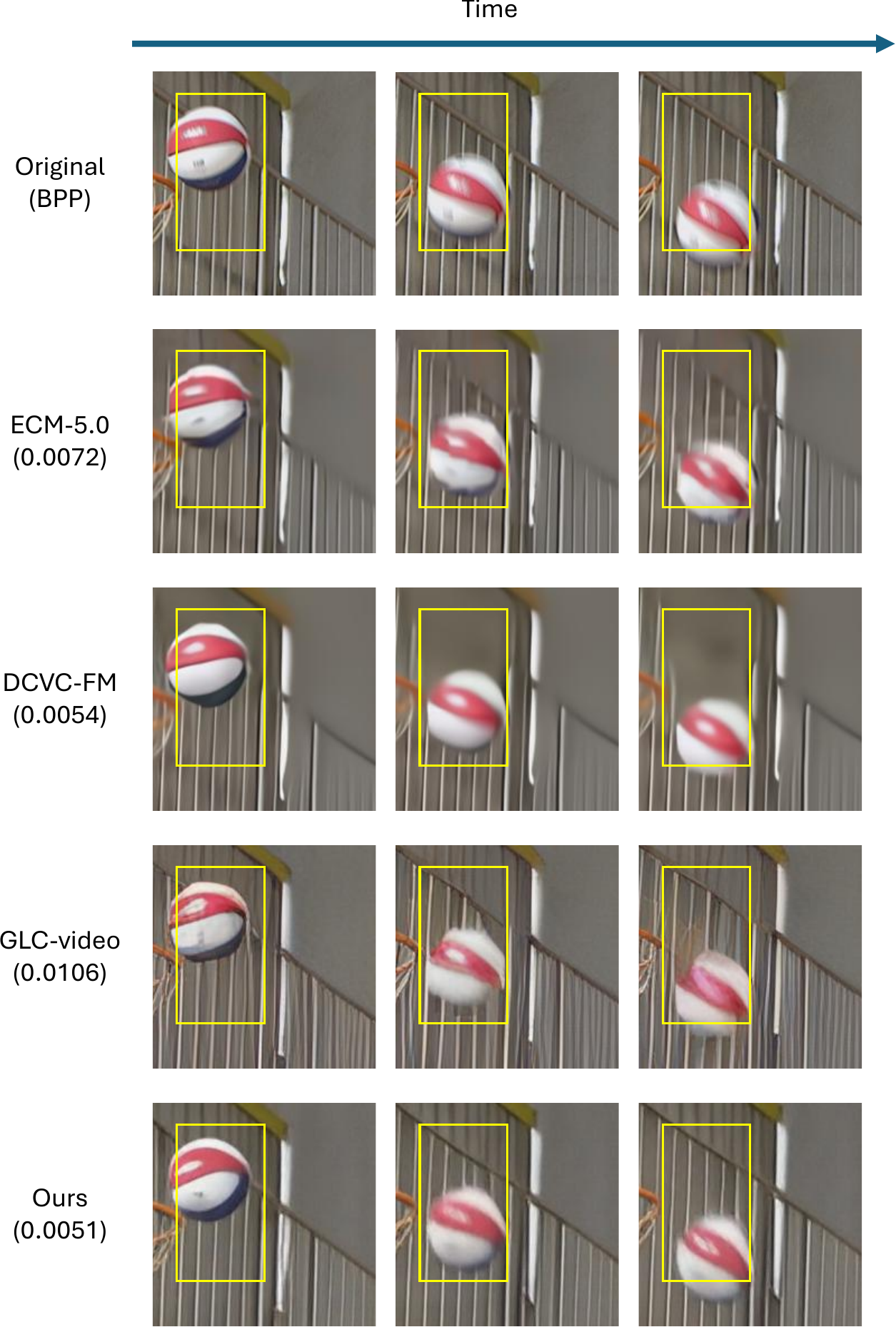}
    \caption{Visualization results across video frames.}
    \label{fig:suppl_time_consistency}
\end{figure*}

For the qualitative evaluation, as illustrated in Fig.~\ref{fig:suppl_time_consistency}, our model preserves the structure of the railing while producing clear and temporally consistent results. In contrast, DCVC-FM and ECM-5.0 produce blurrier reconstructions with noticeable temporal instability, while GLC-video exhibits prominent temporal artifacts across frames. These observations further confirm the ability of our method to generate temporally consistent videos.

\end{document}